%% file: main_arXiv.tex
\documentclass[10pt,twocolumn,letterpaper]{article}

\usepackage{cvpr}           
\usepackage{booktabs}
\usepackage{multirow}
\usepackage{tabularx}
\usepackage{algorithm}
\usepackage{algpseudocode}
\usepackage{graphicx} 
\usepackage{amsfonts} 
\usepackage{amsmath} 
\usepackage{comment}
\usepackage{colortbl}

\input{preamble}

\input{definitions}

\definecolor{cvprblue}{rgb}{0.21,0.49,0.74}
\definecolor{topcolor}{rgb}{1, 0.8, 0.8}
\definecolor{secondcolor}{rgb}{1, 0.87, 0.7} 
\usepackage[pagebackref,breaklinks,colorlinks,allcolors=cvprblue]{hyperref}

\usepackage[capitalize,nameinlink]{cleveref}
\crefdefaultlabelformat{#2\textbf{#1}#3}
\crefname{equation}{}{}
\creflabelformat{equation}{#2\textup{\bf #1}#3}
\crefname{equation}{Eq.}{Eqs.}
\Crefname{equation}{Equation}{Equations}
\crefname{figure}{Fig.}{Figs.}
\Crefname{figure}{Figure}{Figures}
\crefname{table}{Tab.}{Tabs.}
\Crefname{table}{Table}{Tables}
\crefname{section}{Sec.}{Secs.}
\Crefname{section}{Section}{Sections}
\crefname{problem}{Problem}{Problems}
\Crefname{problem}{Problem}{Problems}
\crefname{definition}{Definition}{Definitions}
\Crefname{definition}{Definition}{Definitions}
\crefname{lemma}{Lemma}{Lemmas}
\Crefname{lemma}{Lemma}{Lemmas}
\crefname{theorem}{Thm.}{Thms.}
\Crefname{theorem}{Theorem}{Theorems}
\crefname{remark}{Rmk.}{Rmks.}
\Crefname{remark}{Remark}{Remarks}
\crefname{enumi}{item}{items}
\Crefname{enumi}{Item}{Items}
\crefname{algocf}{Alg.}{Algs.}
\Crefname{algocf}{Algorithm}{Algorithms}
\crefname{assumption}{Asm.}{Asms.}
\Crefname{assumption}{Assumption}{Assumptions}
\crefname{ALC@unique}{line bla}{lines}
\Crefname{ALC@unique}{Line bla}{Lines}

\def\paperID{783}
\def\confName{CVPR}
\def\confYear{2024}

\title{Gear-NeRF: Free-Viewpoint Rendering and Tracking \\ with Motion-aware Spatio-Temporal Sampling }

\author{Xinhang Liu\textsuperscript{\rm 1 \dag}\thanks{Work mainly done when XL was an intern at MERL.} \quad Yu-Wing Tai\textsuperscript{\rm 2} \quad Chi-Keung Tang\textsuperscript{\rm 1 \dag} \\ Pedro Miraldo\textsuperscript{\rm 3} \quad Suhas Lohit\textsuperscript{\rm 3} \quad
   Moitreya Chatterjee\textsuperscript{\rm 3}\\
\textsuperscript{\rm 1}HKUST \quad \textsuperscript{\rm 2}Dartmouth College  \quad \textsuperscript{\rm 3}Mitsubishi Electric Research Laboratories (MERL)\\
{\tt\small xliufe@connect.ust.hk, yu-wing.tai@dartmouth.edu, cktang@cse.ust.hk,}\\
{\tt\small miraldo@merl.com, slohit@merl.com, metro.smiles@gmail.com}}

\usepackage[dvipsnames]{xcolor}
\newcommand{\MC}[1]{\textcolor{magenta}{{\bf MC:}~#1}}

\begin{document}
\maketitle

\begin{abstract}
Extensions of Neural Radiance Fields (NeRFs) to model dynamic scenes have enabled their near photo-realistic, free-viewpoint rendering. Although these methods have shown some potential in creating immersive experiences, two drawbacks limit their ubiquity: (i) a significant reduction in reconstruction quality when the computing budget is limited, and (ii) a lack of semantic understanding of the underlying scenes. To address these issues, we introduce \textbf{Gear-NeRF}, which leverages semantic information from powerful image segmentation models. Our approach presents a principled way for learning a spatio-temporal (4D) semantic embedding, based on which we introduce the concept of \emph{gears} to allow for stratified modeling of dynamic regions of the scene based on the extent of their motion. Such differentiation allows us to adjust the spatio-temporal sampling resolution for each region in proportion to its motion scale, achieving more photo-realistic dynamic novel view synthesis. At the same time, almost for free, our approach enables free-viewpoint tracking of objects of interest  -- a functionality not yet achieved by existing NeRF-based methods. Empirical studies validate the effectiveness of our method, where we achieve state-of-the-art rendering and tracking performance on multiple challenging datasets. The project page is available at: \href{https://merl.com/research/highlights/gear-nerf}{https://merl.com/research/highlights/gear-nerf}.
\end{abstract}
\def\thefootnote{\dag}\footnotetext{XL and CT are supported in part by the Research Grant Council of the Hong Kong SAR grant no. 16201420.}\def\thefootnote{\arabic{footnote}}

\section{Introduction}
\label{sec:intro}
\begin{figure}[t]
\centering
\includegraphics[width=\linewidth]{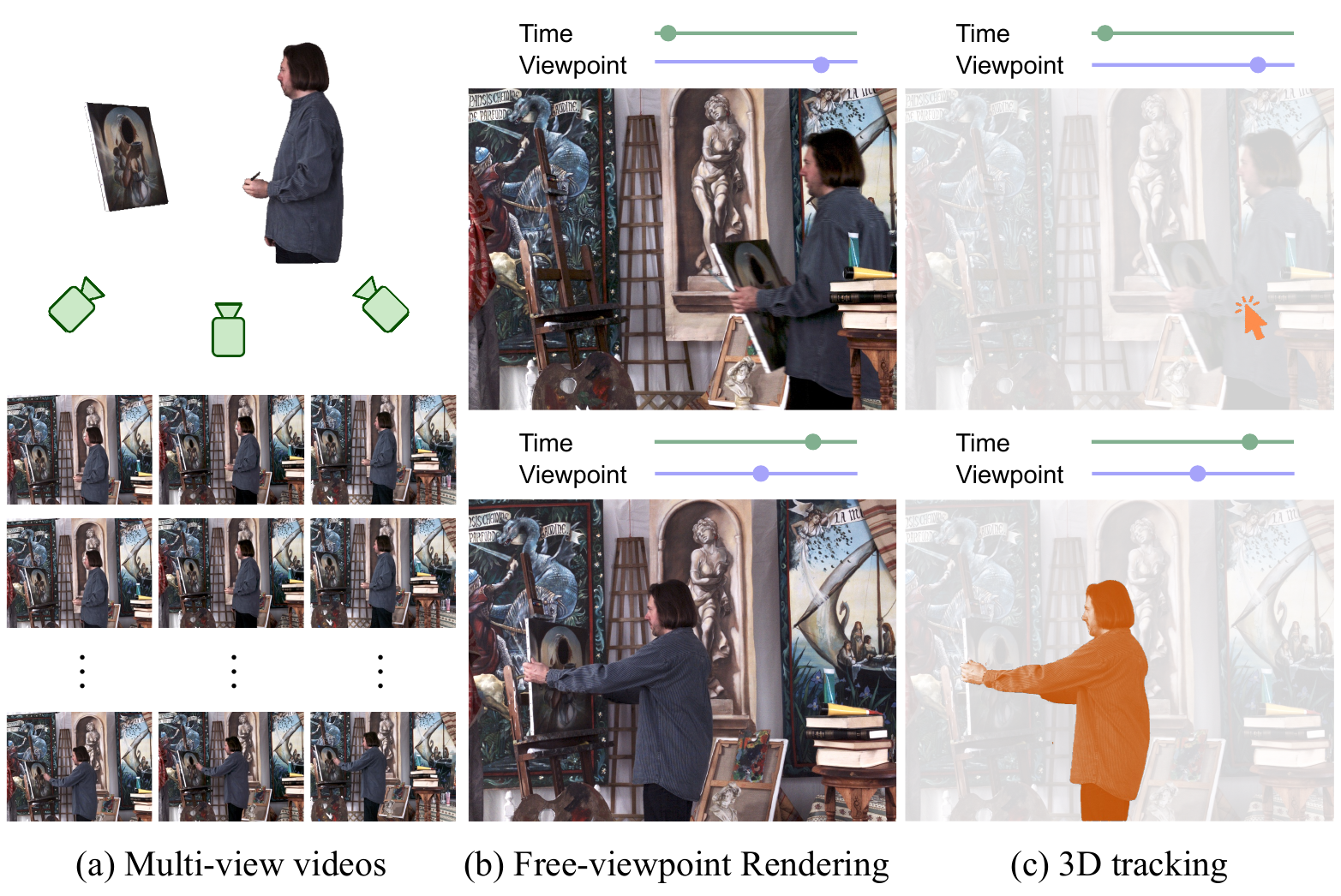}
\vspace{-0.15in}
\caption{\textbf{(a)} Our method takes RGB videos captured from a camera array as input. \textbf{(b)} Trained Gear-NeRF achieves photo-realistic real-time free-viewpoint rendering of a dynamic scene. \textbf{(c)} With users giving a single click at any time and from any viewpoint, our method can perform free-viewpoint tracking of the target object.}
\label{fig:teaser}
\vspace{-0.5cm}
\end{figure}

\begin{figure*}[t]
\centering
\includegraphics[width=\linewidth]{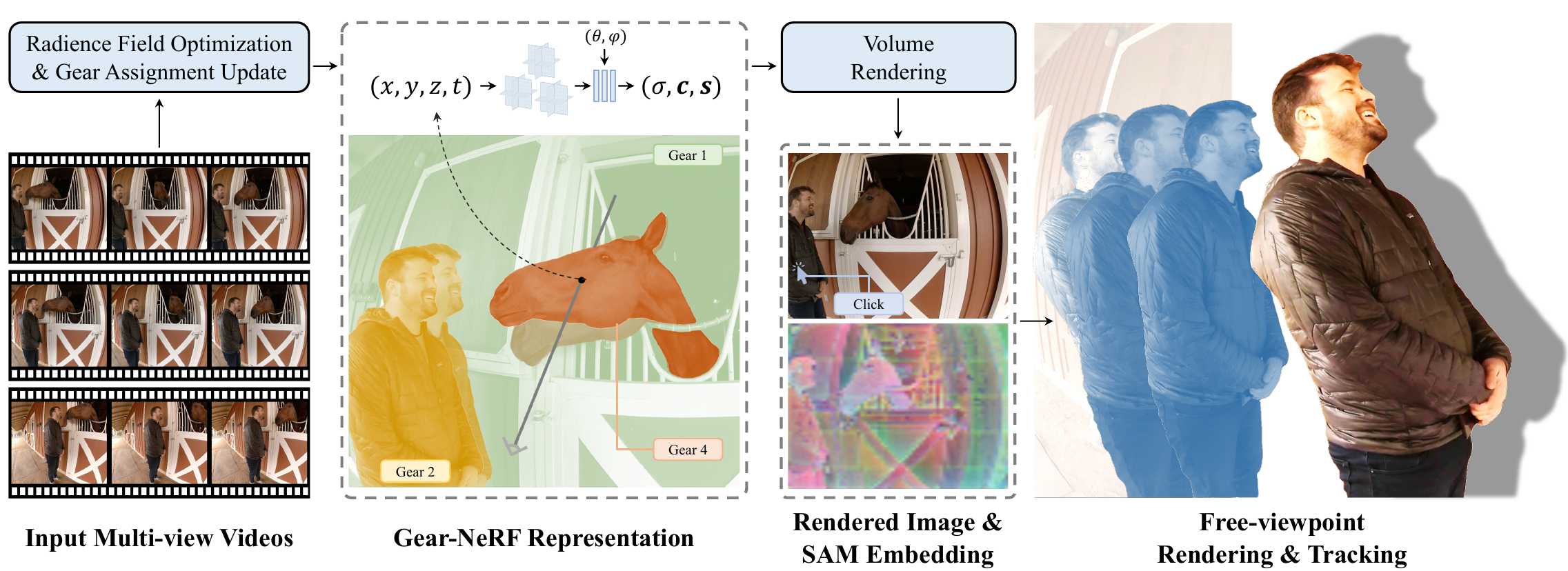}
\caption{\textbf{Pipeline of Gear-NeRF:} Gear-NeRF takes multi-view videos as input. 
After optimizing the serial 4D feature volumes (\Cref{sec:representation}), it maps space-time coordinates to a 4D semantic embedding (\Cref{sec:sam_embedding}), in addition to the volume density and view-dependent radiance color.
Regions with larger motion are automatically assigned higher gear levels (\Cref{sec:gear_determine}) and as a result, receive higher-resolution spatio-temporal sampling (\Cref{sec:st_sampling}).
Furthermore, Gear-NeRF is capable of performing free-viewpoint tracking of a target object with prompts as simple as a user click (\Cref{sec:novel_tracking}).}
\label{fig:pipeline}
\vspace{-0.5cm}
\end{figure*}



Reconstructing 3D scenes has a broad range of applications, including Virtual Reality/Augmented Reality (VR/AR), 3D animation, game production, and film creation which allow users to observe scenes from any desired viewpoint. While it is crucial to reconstruct {\em static} scenes, towards which significant progress has been made, it is even more crucial to reconstruct {\em dynamic} scenes, as the world around us is characterized by a constant state of flux, with many objects in it - in a state of motion.

Recent advances in novel view synthesis, such as Neural Radiance Fields (NeRFs)~\cite{mildenhall2020nerf} have inspired numerous studies to extend them to dynamic 3D scenes. 
Existing approaches either employ a deformation field to map neural fields from a given time to a canonical space~\cite{dnerf, park2021nerfies, stnerf, park2021hypernerf, du2021neural, li2021neural}, or directly model dynamic scenes as a 4D space-time grid~\cite{cao2023hexplane, kplanes, attal2023hyperreel}. 
Though these methods offer improved rendering quality by utilizing more accessible inputs compared to previous solutions~\cite{li2009robust, li2012temporally, taylor2012vitruvian, collet2015high}, they still struggle to ensure rendering quality in low-resource settings, requiring carefully engineered efforts.
Further, most dynamic radiance field approaches adopt a naive spatio-temporal sampling strategy, without discerning the different scales of motion across different regions in the scene. 

We propose to fix this issue by leveraging a semantic understanding of dynamic scenes.
Intuitively, a reconstruction system aware of the distinction between static and dynamic regions in a scene can perform more focused sampling in the dynamic regions, which inherently require more resources per unit volume than static regions, due to their time-evolving nature.
Accordingly, dynamic regions can be further stratified according to their scale of motion. 
To this end, this paper presents \emph{Gear-NeRF}, a framework that leverages semantic embedding from powerful image segmentation models~\cite{sam} for stratified modeling of 4D scenes.
Gear-NeRF optimizes for a 4D semantic embedding, based on which we introduce the concept of  {\it gear} to smartly determine the appropriate region-specific resolution of spatio-temporal sampling in the NeRF.
Regions with larger motion scales are assigned higher gears, through our gear determination scheme and we accordingly perform higher-resolution spatio-temporal sampling.
Empirical studies reveal that this motion-aware sampling strategy improves the quality of synthesized images, over competing approaches.
As a by-product of our semantically embedded representation, we achieve free-viewpoint object tracking, given user prompts. 
\Cref{fig:teaser} presents an overview of the capabilities of our method.

\emph{Gear-NeRF} makes two primary advancements: 
(i) enhanced dynamic novel view synthesis by resorting to smarter spatio-temporal sampling, and (ii) the ability for free-viewpoint tracking of objects of interest. The latter is a capability not yet realized by existing NeRF methods for dynamic scenes. We perform extensive experiments on multiple datasets to validate the generalizability and robustness of our method, which shows state-of-the-art performances for both tasks across all datasets. 

\section{Related Work}
\noindent\textbf{Neural Radiance Fields:}
NeRF~\cite{mildenhall2020nerf} is a recent breakthrough among novel view synthesis methods that uses multilayer perceptrons (MLPs) to parameterize the appearance and density for each point in 3D space, given any viewing direction of the scene. 
Researchers have extended NeRF along various dimensions~\cite{tewari2022advances}, including improving rendering quality~\cite{barron2021mip, barron2022mip, Chen2023ARXIV, chen2023neurbf, hu2023tri, barron2023zip}, handling challenging conditions such as large 
scenes~\cite{tancik2022block, meuleman2023progressively, rematas2022urban}, view-dependent appearances~\cite{verbin2022ref, guo2022nerfren, liu2023clean}, and sparse inputs~\cite{jain2021putting, niemeyer2022regnerf, yang2023freenerf, wynn2023diffusionerf, liu2023deceptive, wu2023reconfusion}. 
NeRF-like neural representations have also found applications in semantic segmentation~\cite{zhi2021place, kobayashi2022decomposing, liu2022unsupervised} and 3D content generation~\cite{chan2022efficient, chan2021pi, poole2022dreamfusion, liu2023zero}. 
Recent work has shown that replacing the deep MLPs with a feature voxel grid can significantly improve training and inference speed~\cite{chen2022tensorf, muller2022instant, sun2022direct, fridovich2022plenoxels}. 
On the other hand,  a more recent approach to further improve visual quality, rendering time, and performance entails representing the scene with 3D Gaussians ~\cite{kerbl20233d}.
Our approach, while drawing upon many of these approaches, deals with dynamic scenes which is beyond the scope of these methods.

\noindent\textbf{Neural Representations for Dynamic Scenes:}
NeRF-like representations have recently been extended to model dynamic scenes in high fidelity~\cite{icsik2023humanrf,lin2022efficient,lin2023im4d,xu20234k4d, li2023dynibar,Chen_2023_CVPR, Geng_2023_CVPR, Yu_2023_CVPR, kirschstein2023nersemble, wang2023neuwigs, peng2023representing, yan2023nerf, liu2023robust, qiao2023dynamic}. 
One straightforward approach to do this is to directly condition the radiance field on time~\cite{gao2021dynamic, li2021neural, li2022neural, xian2021space}. 
Alternatively, several methods model a deformation field to map coordinates from different time stamps to a common canonical space
~\cite{du2021neural, stnerf, park2021nerfies, park2021hypernerf, dnerf, tretschk2021non, yuan2021star, fang2022fast}.
Some recent approaches~\cite{cao2023hexplane, kplanes, wang2022mixed, shao2023tensor4d} represent the scene using a 4D space-time grid, which is decomposed into sets of planar representations for training efficiency.  
Other techniques for improving rendering fidelity and frame rate include Fourier PlenOctrees~\cite{wang2022fourier}, ray-conditioned sample prediction networks~\cite{attal2023hyperreel}, 4D space decomposition (static/dynamic/newly appeared regions)~\cite{song2023nerfplayer}, and explicit voxel grids~\cite{fang2022fast}. 
3D Gaussians have also been adapted to model dynamic scenes~\cite{wu20234d, yang2023real, luiten2023dynamic, yang2023deformable}. 
While these approaches paved the initial path for rendering dynamic scenes, semantically aware modeling of the scene is absent, a caveat that our proposed method seeks to address.

\noindent\textbf{Segment Anything Model (SAM):}
SAM~\cite{sam} is a powerful promptable image segmentation model, which showcases remarkable zero-shot generalization abilities and can produce semantically consistent masks, given a single foreground point on the image. 
HQ-SAM~\cite{ke2023segment} is an improvement on SAM that enhances the quality of masks, especially on objects with intricate boundaries and structures.
Recent works~\cite{yang2023track, cheng2023segment} have extended SAM to perform interactive video object segmentation. These methods utilize SAM for mask initialization or correction and then employ state-of-the-art mask trackers~\cite{cheng2022xmem, yang2022decoupling} for mask tracking and prediction~\cite{chen2023iquery}. Recent methods have also leveraged SAM for tracking multiple reference objects in a video~\cite{zhang2023personalize, chen2023sportsslomo}. 
This work uses SAM to profile the scene into semantic regions, which are then grouped based on motion scales.

\noindent\textbf{3D Semantic Understanding: } 
Existing methods for 3D visual understanding~\cite{huang2018recurrent, tang2022point, yang2019learning, Chen_2020_SketchAwareSSC} mainly focus on closed set segmentation of point clouds or voxels. 
NeRF's ability to integrate information across multiple views has led to its applications in 3D semantic segmentation~\cite{zhi2021place}, object segmentation~\cite{fan2022nerf,liu2022unsupervised,mirzaei2023spin}, panoptic segmentation~\cite{Siddiqui_2023_CVPR}, and interactive segmentation~\cite{isrfgoel2023,nvos}.
%
Kobayashi \etal~\cite{kobayashi2022decomposing} explored the effectiveness of embedding pixel-aligned features~\cite{li2022languagedriven, caron2021emerging} into NeRFs for 3D manipulations.
LERF~\cite{lerf2023} fuses CLIP embeddings~\cite{clip} and NeRFs to enable language-based localization in 3D NeRF scenes. 
Recent work has enabled click/text-based 3D segmentation by learning a 3D SAM embedding~\cite{chen2023interactive} or inverse rendering of SAM-generated masks \cite{sa3d}.
We, on the other hand, seek to utilize the synergy of dynamic NeRFs and SAM segments to derive a semantic understanding of a dynamic 3D scene for tracking objects of interest in novel views -- a first of its kind effort. 

\section{Preliminaries}
\label{sec:preliminary}

\noindent \textbf{Neural Radiance Fields (NeRFs):} Vanilla NeRFs~\cite{mildenhall2020nerf} employ a multi-layer perceptron (MLP) with sinusoidal positional encoding to map a 3D-spatial coordinate $\mathbf{x}=(x,y,z)$ and a viewing direction $\mathbf{d}=(\theta,\phi)$ to a volume density $\sigma \in [0, 1]$ and an emitted RGB, $\mathbf{c}\in\mathbb{R}^3$.
Rendering each image pixel involves casting a ray $\mathbf{r}(t)=\mathbf{o}+t\mathbf{d}$ from the camera center $\mathbf{o}$ through the pixel along direction $\mathbf{d}$. The predicted color for the corresponding pixel is computed as:
\begin{equation}
\label{eqn:volume_rendering}
    \hat{\mathbf{C}}(\mathbf{r}) = \sum_{i=1}^N T_i\alpha_i\mathbf{c}_i, 
\end{equation}
where $T_i=\operatorname{exp}\left(-\sum_{j=1}^{i-1}\sigma_j\delta_j\right)$, $\alpha_i = 1-\exp(-\sigma_i\delta_i)$, and $\delta_j = t_{j+1} - t_j$.
A vanilla NeRF is trained by minimizing the mean squared error between the input images and the predicted images, obtained by rendering the scene from the viewpoints from which the input images have been captured, with the training loss given by:
\begin{equation}
\label{eqn:rgbloss}
    \mathcal{L}_{\text{pho}}=\sum_{\mathbf{r}\in\mathcal{R}}\|\hat{\mathbf{C}}(\mathbf{r)}-\mathbf{C}(\mathbf{r})\|_2^2.
\end{equation}
where $\mathcal{R}$ is the set of all rays projected from the input image. 

\noindent \textbf{Planar-Factorized 4D Volumes:} 
A recent emerging trend of handling dynamics using radiance field representations is to directly adapt them to be conditioned on a frame index $t$ (denoting time) in addition to $\mathbf{x}$ and $\mathbf{d}$. This can be accomplished by learning a mapping from $(\mathbf{x},\mathbf{d}, t)$ to $(\sigma,\mathbf{c})$ using planar-factorized 4D volumes~\cite{cao2023hexplane, kplanes, attal2023hyperreel, shao2023tensor4d}.
These methods attempt to learn a 4D feature vector for every $(\mathbf{x},t)$, by projecting it to a set of 2D-planes. Embeddings of these projections on these planes can then be integrated to obtain the embedding for the 4D point. This can be mathematically represented as follows: 
\begin{align}
\label{eqn:4dvolume}
    \mathbf{f}(\mathbf{x},t) &= \mathbf{B}_1(\mathbf{h}_1(x, y)\odot\mathbf{k}_{1}(z,t)) \nonumber \\
    &+ \mathbf{B}_2(\mathbf{h}_2(x, z)\odot\mathbf{k}_{2}(y,t)) \nonumber \\
    &+ \mathbf{B}_3(\mathbf{h}_3(y, z)\odot\mathbf{k}_{3}(x,t)).
\end{align}
where $\mathbf{h}_i(\cdot, \cdot)$ and $\mathbf{k}_i(\cdot, \cdot)$ are functions (evaluated by bilinear interpolation on regularly spaced 2D feature grids) embedding coordinate tuples to features of dimension $M$,   ``$\odot$'' denotes an element-wise 
product, and $\mathbf{B}_i(\cdot)$ denotes a linear transform which maps the products to feature vectors.
Subsequently, a tiny MLP can map the feature vector $\mathbf{f}(\cdot, \cdot)$ to the volume density, $\sigma$, and the view-dependent emitted color, $\mathbf{c}$, given the viewing direction $\mathbf{d}$. 


\section{Proposed Method}

Given a set of  $W$ input videos, $\mathcal{V} = \set{\vid_1,\vid_2,\cdots, \vid_W}$ of a dynamic scene, with calibrated camera poses, our approach represents the scene using a series of 4D feature volumes (\Cref{sec:representation}) along with 4D semantic embeddings (\Cref{sec:sam_embedding}).

Analogous to multiple gears in motor vehicles for optimizing engine performance, Gear-NeRF stratifies this semantically embedded scene representation into $N_{\text{gear}}$ levels, based on the motion scales. Each of these levels is called a \emph{gear}.
Through our training scheme, regions with larger motion are assigned higher gear levels (\Cref{sec:gear_determine}) and as a result, are more densely sampled (\Cref{sec:st_sampling}) for improved dynamic novel view synthesis.
Our 4D semantic embedding also enables a new functionality, almost for free -- free-viewpoint tracking of target objects, given simple user prompts like clicks (\Cref{{sec:novel_tracking}}). \Cref{fig:pipeline} shows the overall pipeline of \emph{Gear-NeRF}.

\subsection{Serial 4D Feature Volumes}
\label{sec:representation}
Instead of using a unified 4D volume to represent a dynamic scene~\cite{cao2023hexplane, kplanes, attal2023hyperreel, shao2023tensor4d}, our representation consists of a series of feature volumes, each corresponding to a gear level, $\mathcal{G}$. 
Specifically, for any space-time coordinate $(\mathbf{x}, t)$, its feature vector corresponding to $\mathcal{G}$ is computed as follows:
\begin{align}
\label{eqn:geared_volume}
    \mathbf{f}^{\mathcal{G}}(\mathbf{x},t) &= \mathbf{B}_1(\mathbf{h}_1(x, y)\odot\mathbf{k}_{1}^{\mathcal{G}}(z,t)) \nonumber \\
    &+ \mathbf{B}_2(\mathbf{h}_2(x, z)\odot\mathbf{k}_{2}^{\mathcal{G}}(y,t)) \nonumber \\
    &+ \mathbf{B}_3(\mathbf{h}_3(y, z)\odot\mathbf{k}_{3}^{\mathcal{G}}(x,t)).
\end{align}
The vector-valued functions $\mathbf{h}_j(\cdot, \cdot)$ and linear transforms $\mathbf{B}_j(\cdot)$ are shared by all gears, while each gear has its own spatio-temporal embedding $\mathbf{k}_{j}^{\mathcal{G}}(\cdot, \cdot)$, in $M$-dimensional space. 
Therefore, each gear describes regions of a certain scale of motion while the purely spatial features can be shared among all gears.

We obtain the gear level at any spatio-temporal coordinate also from a planar-factorized 4D feature volume. 
Specifically, the gear level at $(\mathbf{x}, t)$ is computed as: 
\begin{align}
\label{eqn:gear_assign}
        g(\mathbf{x},t) &= \mathbf{1}^\top(\mathbf{h}_{1}'(x, y)\odot\mathbf{k}_{1}'(z,t)) \nonumber\\ 
        &+ \mathbf{1}^\top(\mathbf{h}_{2}'(x, z)\odot\mathbf{k}_{2}'(y,t)) \nonumber\\ 
        &+ \mathbf{1}^\top(\mathbf{h}_{3}'(y, z)\odot\mathbf{k}_{3}'(x,t)),
\end{align}
where $\mathbf{1}$ is a vector of ones, $\mathbf{h}_i' (\cdot, \cdot)$ and $\mathbf{k}_i' (\cdot, \cdot)$ are $M$-dimensional embedding functions. 
This however defines a continuous feature volume. To map it to the gear level integers, we apply the following projection operation:
\begin{align}
\label{eqn:round}
        \begin{array}{ll} p(\mathbf{x},t) & = \begin{cases} 1, & \text{if } g(\mathbf{x},t) < 1, \\ N_{\text{gear}}, & \text{if } g(\mathbf{x},t) \geq N_{\text{gear}}, \\ 
        \left\lceil g(\mathbf{x},t) \right\rceil, & \text{otherwise.}  
        \end{cases} \end{array}
\end{align}
Based on this gear level volume, we define a 4D mask for a region at gear level $\mathcal{G}$ as:
\begin{equation}
\begin{array}{ll} m_{\mathcal{G}}(\mathbf{x},t) & = \begin{cases} 1, & \text{if } p(\mathbf{x},t) = \mathcal{G}, \\ 0, & \text{otherwise}. \end{cases} \end{array}
\end{equation}
The final feature vector at $(\mathbf{x}, t)$ is computed as:
\begin{equation}
    \mathbf{f}(\mathbf{x},t)=\sum_{\mathcal{G}=1}^{N_{\text{gear}}} m_{\mathcal{G}}(\mathbf{x},t)\mathbf{f}^{\mathcal{G}}(\mathbf{x},t).
\end{equation}
Subsequently, a tiny MLP, $F_\theta$, maps these feature vectors $\mathbf{f}(\cdot, \cdot)$ as well as the viewing direction $\mathbf{d}$ to the volume density $\sigma$ and radiance color $\mathbf{c}$. This allows us to obtain a photometric rendering of the scene. 

\subsection{4D Semantic Embedding}
\label{sec:sam_embedding}
\emph{Gear-NeRF} leverages the strong object priors of the SAM~\cite{sam} model to acquire a semantic understanding of the scene, for improved photometric rendering (\Cref{sec:gear_determine} and \Cref{sec:st_sampling}) as well as free-viewpoint object tracking (\Cref{sec:novel_tracking}).
%
Toward this end, we utilize the SAM encoder to obtain 2D feature maps from the frames of each video.
We then optimize a 4D SAM embedding field by supervising it with these 2D feature maps. 
In particular, the MLP above, $F_\theta$, is configured to output a 4D semantic embedding $\mathbf{s}$ for a given space-time coordinate in addition to the density, $\sigma$, and color, $\mathbf{c}$. 
%
To render 2D semantic feature maps in a given view, we compute the semantic feature of a pixel in the feature map by tracing a ray through it and perform volume rendering analogous to \Cref{eqn:volume_rendering}, as follows:
\begin{equation}
    \hat{\mathbf{S}}(\mathbf{r}) = \sum_{i=1}^N T_i\alpha_i\mathbf{s}_i.
\end{equation}
%
This SAM embedding is supervised by minimizing the mean squared error between the prediction and the ground truth features ($\mathbf{S}(\mathbf{r})$) from the SAM encoder, as shown: 
\begin{equation}
\label{eqn:samloss}
\mathcal{L}_{\text{SAM}}=\sum_{\mathbf{r}\in\mathcal{R}}\|\hat{\mathbf{S}}(\mathbf{r)}-\mathbf{S}(\mathbf{r})\|_2^2.
\end{equation}

\begin{figure}[t]
\vspace{-0.15in}
\centering
\includegraphics[width=0.97\linewidth]{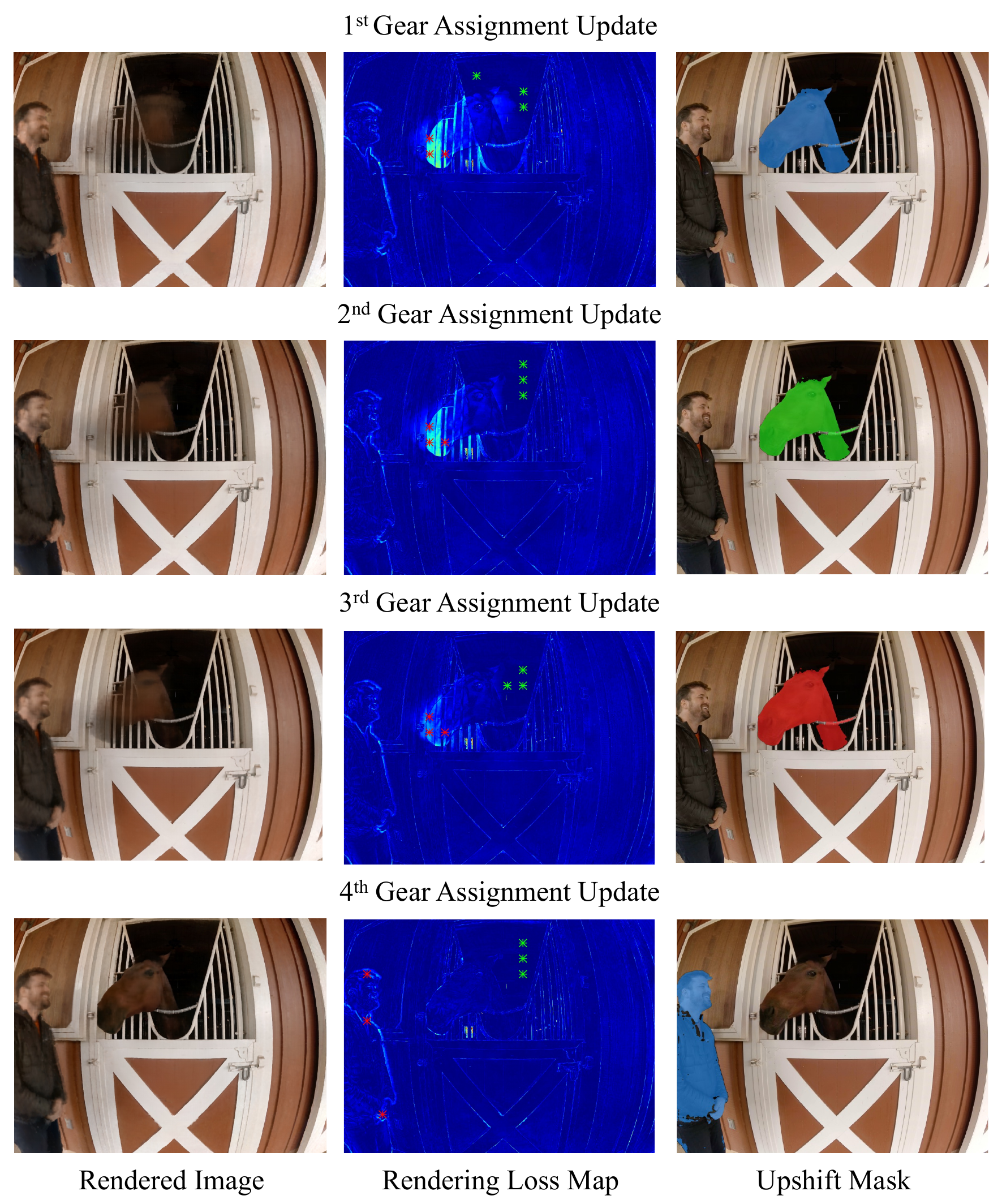}
\caption{\textbf{Illustration of Gear Assignment Update:}  For each gear assignment update, we calculate the \textbf{rendering loss map} between the rendered RGB-SAM map and the ground truth and identify the centers of the patches with the maximum and minimum losses, marked in red and green (second column). These points are then fed into the SAM decoder as positive and negative prompts to generate an \textbf{upshift mask} representing the areas that need to be shifted to a higher gear (last column).
After the first gear assignment update, we see that the next candidate region for upshift is situated where the horse is located, and so on.
Upshift mask colors imply the gear levels after the update (blue-2, green-3, red-4).}
\label{fig:gear_determination}
\vspace{-0.5cm}
\end{figure}

\subsection{Training Scheme with Gear Assignment}
\label{sec:gear_determine}
With gear initialization $g(\mathbf{x},t) = 1, \forall \mathbf{x}, t$, the (semantically embedded) radiance field optimization and gear assignment updating take place in an alternating fashion.

\noindent\textbf{Gear Assignment Update:} 
As illustrated in \Cref{fig:gear_determination}, when updating the gear assignment after a period of radiance field optimization, we find the regions rendered most poorly from the rendering loss maps and increment their gears for denser spatio-temporal sampling.
%
%
%
%
The following steps lay out the process for updating gear assignments to regions:
 \begin{itemize}
     \item We sample a number of viewpoints and time steps and render 2D-images/SAM features for it. For every rendered RGB-SAM map, we compute a rendering loss map. Each pixel of the rendering loss map is computed as: $\mathcal{L}(\mathbf{r}) := \mathcal{L}_{\text{pho}} (\mathbf{r}) + \lambda \mathcal{L}_{\text{SAM}} (\mathbf{r})$. See \Cref{fig:gear_determination} for example loss maps.
     \item Next, we patchify each rendering loss map to find patches with the top-$k$ largest/smallest average loss. The center coordinate of these patches serve as positive/negative prompts for the next step. See \Cref{fig:gear_determination} for example positive (red) / negative (green) prompts.
     \item We then feed the ground truth RGB images together with positive and negative prompts into the SAM decoder~\cite{sam} to estimate an \emph{upshift mask}. These masks tend to cover regions that have motions and are not satisfactorily rendered with the current sampling resolution. We have multiple upshift masks at different viewpoints and time steps.
     \item For every pixel of an upshift mask, we trace a ray and sample points along it and
     update the gear assignment by pushing $g(\mathbf{x},t)$ towards incremented values. 
      \end{itemize}

In particular, in the last step, for each pixel within an upshift mask, a corresponding ray is traced that connects it with the camera center $\mathbf{o}$, along direction $\mathbf{d}$. Next, a set of points are sampled along this ray. The collection of sampled points, that lie on the rays emanating from within the masked region constitutes the set $\mathcal{S}_{\text{upshift}} = \{ (\mathbf{x}^{\text{upshift}}_i, t^{\text{upshift}}_i) \}_{i=1}^{N_{\text{upshift}}}$, where $N_{\text{upshift}}$ represents the total count of points sampled from rays pertaining to the masked region. We then follow a similar procedure to sample a set of points pertaining to the unmasked region. We denote this set as:  $\mathcal{S}_{\text{stay}} = \{ (\mathbf{x}^{\text{stay}}_i, t^{\text{stay}}_i) \}_{i=1}^{N_{\text{stay}}}$, with $N_{\text{stay}}$ indicating the total number of points sampled from rays that pertain to the unmasked area.
Next, for each sample point in each of $\mathcal{S}_{\text{upshift}}$ and $\mathcal{S}_{\text{stay}}$, we query its current gear level $p(\mathbf{x}, t)$. In order to assign new gear levels, we need to update the gear assignment function $g(\cdot,\cdot)$, which proceeds with the objective function: 
\begin{equation}
\begin{aligned}
\mathcal{L}_{\text{upshift}} &= \frac{1}{N_{\text{upshift}}}\sum_{(\mathbf{x}, t)\in\mathcal{S}_{\text{upshift}}} \|g(\mathbf{x}, t;\mathbf{\Theta})-(p(\mathbf{x}, t)+1)\|_2^2 \\
&+ \frac{\lambda_{\text{stay}}}{N_{\text{stay}}}\sum_{(\mathbf{x}, t)\in\mathcal{S}_{\text{stay}}} \|g(\mathbf{x}, t;\mathbf{\Theta})-p(\mathbf{x}, t)\|_2^2,
\end{aligned}
\end{equation}
where $\mathbf{\Theta}$ denotes the set of optimizable parameters for $g(\cdot,\cdot)$. 
The minimization of the first term encourages sample points within the masked region \(\mathcal{S}_{\text{upshift}}\) to have a gear level equal to their current gear incremented by one, resulting in an upshift of gear. Conversely, minimizing the second term encourages the points in \(\mathcal{S}_{\text{stay}}\) to maintain their gear values, thereby encouraging the remaining regions to keep their current gear levels and avoiding unwanted upshifts.
We update $\mathbf{\Theta}$ via a single step of gradient descent as follows:
\begin{equation}
\label{eqn:gear_update}
\mathbf{\Theta} \leftarrow \mathbf{\Theta} - \alpha \nabla_{\mathbf{\Theta}} \mathcal{L}_{\text{upshift}},
\end{equation}
where $\alpha$ is the learning rate. 
Since, $g(\cdot,\cdot)$ is essentially derived from the embedding functions, $\mathbf{h}_i' (\cdot, \cdot)$ and $\mathbf{k}_i' (\cdot, \cdot)$ for $i \in \{1, 2, 3\}$, the aforementioned optimization step amounts to updating these embedding functions. Once updated, we proceed with a fresh round of gear assignment to increase the spatio-temporal sampling resolution for the regions that end up at a higher gear level than before.

\noindent\textbf{Radiance Field Optimization:} With the updated gear assignment, we increase the resolution of spatio-temporal sampling for the gear-shifted regions (\Cref{sec:st_sampling}) and then resume optimizing the radiance field.
We alternate between the two processes: radiance field optimization (each time for $L$ epochs), and gear assignment updates until the average variance of each rendering loss map is below a predetermined threshold. After this, we optimize the radiance field for an additional $L'$  epochs without further gear assignment updates.

\subsection{Motion-aware Spatio-Temporal Sampling}
\label{sec:st_sampling}
In this subsection, we explain our motion-aware spatio-temporal sampling strategy based on assigned gears,  permitting differential processing of regions at different gear levels.
By temporal sampling, we imply the choice of temporal resolution for planar-factorized 4D feature volumes, and by spatial sampling, we mean the strategy used to choose sampling points along each ray for volume rendering.

\noindent\textbf{Motion-aware Temporal Sampling:} To handle the increasing intensity of object motion, as reflected by their growing gear levels, we increment the temporal resolution for voxel grids. 
%
%
Specifically, $\mathbf{k}_{j}^{\mathcal{G}}$  in \Cref{eqn:geared_volume} has increasing resolution along the time axis, thereby empowering the 4D feature volumes to better model the dynamics along the temporal axis. This ensures fast-moving objects can be more faithfully modeled without unsightly blurring. 
The temporal resolution for each gear's feature volume is determined by linear interpolation between $1$ (for $\mathcal{G}=1$) and the total number of frames (for $\mathcal{G}=N_{\text{gear}}$).

\begin{figure}[t]
\centering
\includegraphics[width=0.9\linewidth]{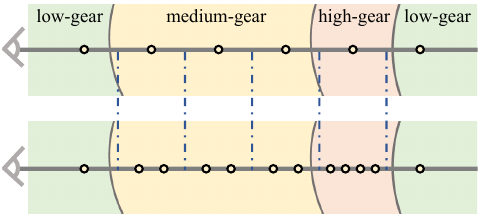}
\caption{\textbf{Motion-aware Spatial Sampling:} We split each sampled point into $2 ^ {p(\mathbf{x}, t)} $ points, equally spaced within the corresponding ray segment. The top row shows the vanilla uniformly sampled points, while the bottom one shows the densely sampled points after splitting using our strategy. 
}
\label{fig:splitting}
\vspace{-0.5cm}
\end{figure}

\noindent\textbf{Motion-aware Spatial Sampling:} While denser sampling of points can improve reconstruction accuracy, increasing the number of sampling points throughout the scene can lead to prohibitive computational costs. Therefore, we propose a 3D point-splitting strategy as illustrated in \Cref{fig:splitting}. We begin by sampling a relatively small number, $n$, of samples along each ray, assuming it is at the lowest gear level. 
If a sampled point belongs to a region with a higher gear, as determined by $p(\mathbf{x}, t)$, we then sample more densely in that region. For every sampled point in that region, we split it into $2 ^ {p(\mathbf{x}, t)-1} $ points, equally spaced within the corresponding ray segment (at that gear level).

\begin{figure*}[t]
\centering
\includegraphics[width=0.9\linewidth]{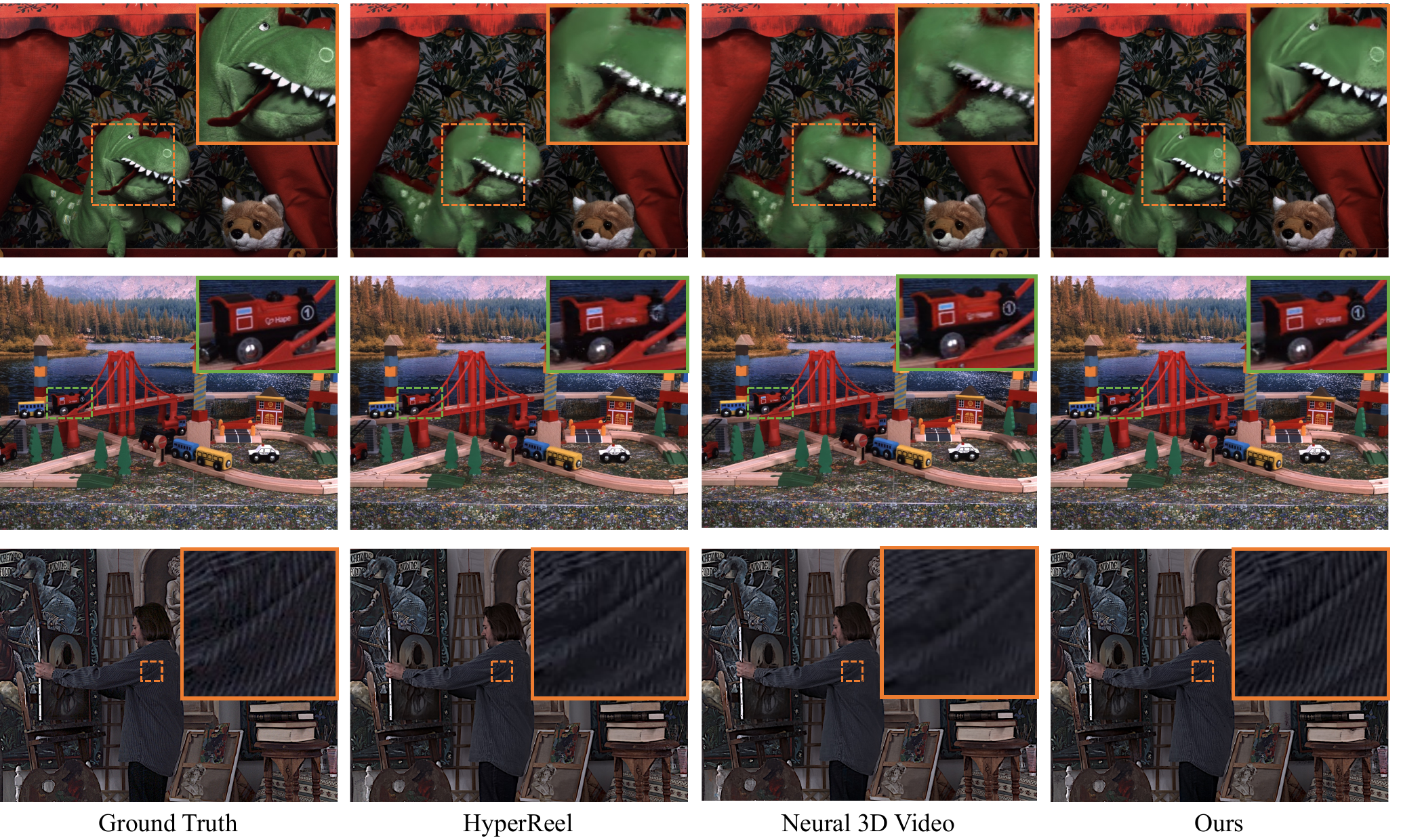}
\caption{\textbf{Qualitative comparisons for novel view synthesis on the Technicolor dataset~\cite{technicolor}: } We qualitatively compare our approach against HyperReel~\cite{attal2023hyperreel} and Neural 3D Video~\cite{li2022neural}. Our approach better recovers fine details like patterns on the toys or stripes on the shirt.}
\label{fig:technicolor}
\vspace{-0.5cm}
\end{figure*}

\subsection{Free-Viewpoint Tracking with User Prompts}
\label{sec:novel_tracking}
Our 4D SAM embedding enables another useful functionality, almost for free --  free-viewpoint object tracking, where the user only needs to provide as few as one click to extract the target object based on the 4D embedding.
Next, we describe how, given a user-supplied point click at any arbitrary viewpoint and time step, we obtain the corresponding object mask at a novel viewpoint and time step.
%

\noindent\textbf{Masks for Novel Viewpoints: }The first step for this task entails finding the 3D correspondence of the user click. 
We trace a ray through the selected pixel, and by utilizing the volume density, we determine the depth at which the ray intersects with the first object surface it encounters. This yields the 3D coordinates of the point of intersection.
Subsequently, the 3D coordinates of this intersection can be easily mapped into a 2D coordinate within any novel viewpoint image, using the camera pose of the new viewpoint.
Alongside the rendered SAM feature map of the novel view, we feed this coordinate into the SAM decoder to generate the object mask for the novel view.

\noindent\textbf{Masks for Novel Time Steps: }For this task, we propagate an object mask to its neighboring time step.
Specifically, with an object mask for a specific frame $t$, we calculate the bounding box of this mask and use this bounding box as a prompt to SAM for neighboring frames $t'=t+1$ or $t' =t-1$.
By inputting this prompt along with the rendered SAM feature map at $t'$ into the SAM decoder, we can obtain the object mask for $t'$. 
Combining the above two processes, we can start from a single click and get the object mask in any viewpoint and time step.

\section{Experiments}
We assess the performance of our proposed Gear-NeRF for dynamic novel view synthesis and free-viewpoint object tracking across a range of challenging datasets, comparing it with state-of-the-art methods. Through ablation studies, we provide empirical evidence of the effectiveness of its fundamental components. 
We kindly refer the reader to our supplementary material for additional experimental details and results, including videos for free-viewpoint rendering and tracking.

\subsection{Experimental Setup}
\noindent\textbf{Implementation Details: }We implement our method using PyTorch \cite{paszke2019pytorch} and conduct experiments on an NVIDIA RTX 4090 GPU with 24 GB RAM. We divide each input video into chunks of 100 frames. For every chunk, we train a model for approximately 2.5 hours. Our 4D feature volumes yield embeddings with a dimension of $M = 32$. We set the gear number $ N_{\text{gear}} $ to $4$.
We find patches with top-$k=3$ largest/smallest average loss for gear assignment updates to obtain prompts. The rendering loss map is computed with $\lambda=0.01$.
For the optimization of radiance fields, $L=3$ and $L'=10$.
In our motion-aware spatial sampling, each ray initially has $n=64$ sampling points. We use an initial learning rate of 0.02 for all the parameters and optimize them using ADAM~\cite{kingma2014adam}. For \Cref{eqn:gear_update}, we use $\alpha=0.02$.

\noindent\textbf{Datasets: }
\textbf{(i) The Technicolor light field dataset} \cite{technicolor} includes diverse indoor environment videos captured by a 4$\times$4 camera rig. We evaluate on 4 sequences (\emph{Train, Theater, Painter, Birthday}) at the original 2048$\times$1088 resolution, holding out the same view as prior work~\cite{attal2023hyperreel} (the second row and second column) for evaluation. 
\textbf{(ii) The Neural 3D Video dataset} \cite{li2022neural} includes indoor multi-view video sequences captured by 20 cameras at a resolution of 2704$\times$2028 pixels. We experiment on 6 sequences (\emph{Cut Roasted Beef, Flame Steak, Coffee Martini, Cook Spinach, Flame Salmon, Sear Steak}), downsampling by a factor of 2 and holding out the central view (akin to prior work~\cite{attal2023hyperreel}) for evaluation. 
\textbf{(iii) The Google Immersive dataset}\cite{broxton2020immersive} contains light field videos of indoor and outdoor scenes captured by a time-synchronized 46-fisheye camera rig, with a resolution of 2560$\times$1920 pixels. We experiment with 9 sequences from it (\emph{Flames, Truck, Horse, Car, Welder, Exhibit, Face Paint 1, Face Paint 2, Cave}). We downsample the video resolution by a factor of 2  and hold out the central view (like prior work~\cite{attal2023hyperreel}) for evaluation.
For our experiments, we adopted the same resolution and held-out view selection as prior work~\cite{attal2023hyperreel}.

\noindent\textbf{Evaluation Metrics:} 
For the task of novel view synthesis of dynamic scenes, we evaluate using the following standard metrics: (i) Peak Signal-to-Noise Ratio (PSNR), (ii) Structural Similarity Index Measure (SSIM)~\cite{wang2004image} and (iii) Learned Perceptual Image Patch Similarity (LPIPS)~\cite{zhang2018unreasonable}, by comparing the reconstructed frames against ground truth images. These metrics are computed on the held-out view and averaged across all frames.
For the free-viewpoint object tracking task, we designate a specific viewpoint and time step for the user to give prompting clicks. Subsequently, we assess the quality of the predicted object masks at novel viewpoints. The quality of the object mask is quantified in terms of the Mean Intersection over Union (mIoU) and accuracy (Acc.), which are calculated against the ground truth mask, manually annotated utilizing Adobe Photoshop. Additionally, we present the same metrics computed for \emph{novel time steps}, denoted as t-mIoU and t-Acc.

\noindent\textbf{Baselines:}
We run a comprehensive comparison of our method against a range of recent NeRF-based baseline methods: (i) ST-NeRF~\cite{stnerf}, (ii) HexPlane~\cite{cao2023hexplane}, (iii) HyperReel~\cite{attal2023hyperreel}, and (iv) MixVoxels~\cite{wang2022mixed}.
As the first method to enable promptable free-viewpoint object tracking under the NeRF setting, there is no established direct baseline for this specific task. However, SA3D~\cite{sa3d}, a recent method for segmenting static scenes, is treated as a baseline for predicting masks at novel viewpoints corresponding to the \emph{prompted time step}.

\subsection{Results}
\noindent\textbf{Dynamic Novel View Synthesis:}
As shown in \Cref{fig:technicolor}, Gear-NeRF produces high-quality novel view synthesis of dynamic scenes, accurately modeling real-world scenes with intricate motions and fine details. For example, patterns on the toys or stripes on the shirt, are faithfully rendered, resulting in more photo-realistic images compared to all of the baselines. \Cref{tab:results}, presents quantitative comparisons of our method against the baselines.
While Gear-NeRF has longer training (Tr. Time) / inference times (FPS) compared to some baselines, it almost always achieves the best performance in terms of rendering quality.
%
%

\begin{figure*}[t]
\centering
\includegraphics[width=0.9\linewidth]{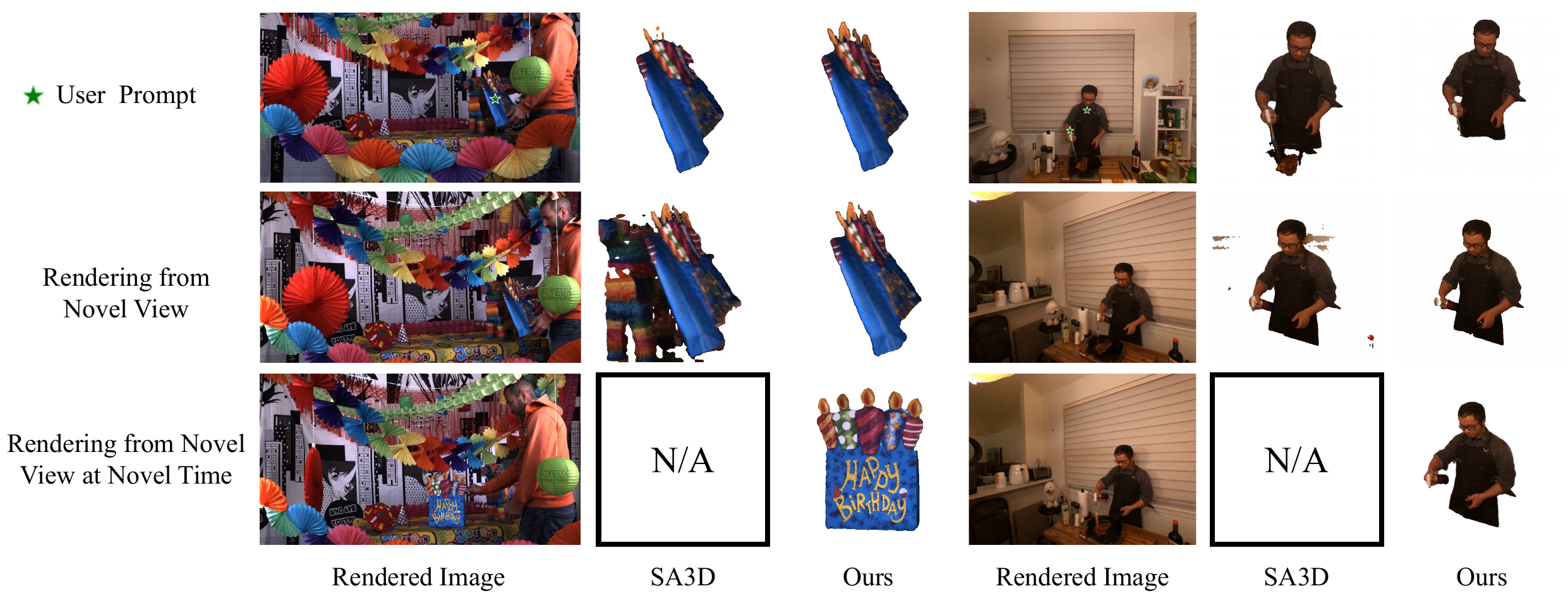}
\caption{\textbf{Qualitative comparisons of free-viewpoint object tracking on Technicolor~\cite{technicolor} and Neural 3D Video~\cite{li2022neural} datasets: } 
Our method can obtain desirable object masks, with clear edges, from prompting points provided by users at desired time steps and viewpoints. 
}
\label{fig:tracking_result}
\vspace{-0.5cm}
\end{figure*}

\begin{table}[t]
\centering
\caption{\textbf{Quantitative comparisons for dynamic novel view synthesis:} Our method outperforms all baselines across all datasets on all metrics. We report means over all scenes for each dataset. \textbf{Best} and \underline{second best} results are highlighted.}
\vspace{-0.3cm}
\label{tab:results}
\resizebox{\linewidth}{!}{%
\setlength{\tabcolsep}{5pt}%
\begin{NiceTabular}{@{}llccccc@{}}[code-before =%
\rectanglecolor{gray!20}{5-2}{5-7}%
\rectanglecolor{gray!20}{9-2}{9-7}%
\rectanglecolor{gray!20}{13-2}{13-7}%
]
\toprule
\thead[l]{Dataset} & \thead[l]{Method} & \thead[c]{PSNR ($\uparrow$)} & \thead[c]{SSIM ($\uparrow$)} & \thead[c]{LPIPS ($\downarrow$)}  & {\thead[c]{Tr. Time($\downarrow$)}} & {\thead[c]{FPS($\uparrow$)}} \\ 
\midrule
\multirow{4}{*}{Technicolor~\cite{technicolor}}
 & ST-NeRF~\cite{stnerf} & 30.86 & 0.883 & 0.101 & 344 min & 0.7 \\
 & HyperReel~\cite{attal2023hyperreel} & \ul{31.04} & \ul{0.887} & \ul{0.092} & \ul{97 min} & 7.7  \\
 & MixVoxels~\cite{wang2022mixed} &28.99 	&0.842  &  0.103	 &\textbf{14 min} & \textbf{18.9} \\
 & Ours & \textbf{32.21}	& \textbf{0.919} &	\textbf{0.058} & 148 min & \ul{7.9}\\
 \midrule
\multirow{4}{*}{Google Immersive~\cite{broxton2020immersive}} 
  & HexPlane~\cite{cao2023hexplane} &27.67 &	0.808 &	0.188 & 527 min & 0.9 \\
 & HyperReel~\cite{attal2023hyperreel} &\ul{28.32} &	\ul{0.862} &	\ul{0.145}  & \ul{117 min} & \ul{7.4}\\
 & MixVoxels~\cite{wang2022mixed} & 27.14	& 0.835 &	0.209 &\textbf{19 min} & \textbf{16.8}\\
 & Ours & \textbf{28.74} &	\textbf{0.876} &	\textbf{0.122}  & 189 min & 7.0  \\
\midrule
\multirow{4}{*}{Neural 3D Video~\cite{li2022neural} }
 & ST-NeRF~\cite{stnerf} & 31.03 & 0.890 & 0.081 & 367 min & 0.7 \\
 & HyperReel~\cite{attal2023hyperreel} & \ul{31.12} & 0.928 & \ul{0.065} & \ul{129 min} & 6.1  \\
 & MixVoxels~\cite{wang2022mixed} &30.69 	& \textbf{0.944} &	0.139 & \textbf{15 min} & \textbf{16.3} \\
 & Ours & \textbf{31.80} & \ul{0.936} & \textbf{0.058} &204 min &\ul{ 6.8} \\
\bottomrule
\end{NiceTabular}%
}
\vspace{-0.7cm}
\end{table}

\noindent\textbf{Free-Viewpoint Tracking: }
In \Cref{fig:tracking_result}, we present the object masks obtained by our method based on the user prompts provided at a specified viewpoint and time step. Specifically, the first row displays masked objects at the prompted viewpoint and time step. The second row shows novel view masks at the prompted time step. The third row shows novel view masks at novel time steps.
We see that masks obtained from Gear-NeRF show precise boundaries, compared to SA3D.
\Cref{tab:tracking} presents quantitative assessment of the quality of the masks generated of our method versus SA3D. Our method exceeds 90\% across all metrics and datasets, demonstrating the effectiveness of our approach.
%
Our gains over SA3D can be attributed to the fact that SA3D, as a static scene segmentation method, does not utilize information across all time frames, whereas our approach does. 
SA3D is incapable of predicting masks for novel time steps, and as such, the corresponding entries are marked as not applicable, a shortcoming which our method does not have.


\begin{table}[t]
\centering
\caption{\textbf{Quantitative comparisons for free-viewpoint tracking:} t-mIoU and t-Acc are metrics used for evaluating novel view masks at novel time steps, not applicable to SA3D. Reported metrics are averages over all scenes for each dataset.}
\vspace{-0.3cm}
\label{tab:tracking}
\resizebox{1\linewidth}{!}{%
\setlength{\tabcolsep}{5pt}%
\begin{NiceTabular}{@{}llcccc@{}}[code-before =%
\rectanglecolor{gray!20}{3-2}{3-6}%
\rectanglecolor{gray!20}{5-2}{5-6}%
\rectanglecolor{gray!20}{7-2}{7-6}%
]
\toprule
\thead[l]{Dataset} & \thead[l]{Method} & \thead[c]{mIoU ($\uparrow$)} & \thead[c]{Acc. ($\uparrow$)} & \thead[c]{t-mIoU ($\uparrow$)} & \thead[c]{t-Acc. ($\uparrow$)} \\ 
\midrule
\multirow{2}{*}{Technicolor~\cite{technicolor}}
 & SA3D~\cite{sa3d} & 96.4	& 97.1& N/A &N/A \\
 & Ours &\textbf{ 97.4}	 &\textbf{97.6} &	\textbf{92.1} &	\textbf{93.3 }\\
\midrule
\multirow{2}{*}{Google Immersive~\cite{broxton2020immersive}}
 & SA3D~\cite{sa3d} &94.1	&94.8	&N/A	&N/A\\
 & Ours & \textbf{94.3}&	\textbf{95.0}&	\textbf{91.5}&	\textbf{92.8} \\
 \midrule
\multirow{2}{*}{Neural 3D Video~\cite{li2022neural}}
 & SA3D~\cite{sa3d} & 93.1 &	94.0 &	N/A &	N/A\\
 & Ours &\textbf{93.4}	&\textbf{94.3}	&\textbf{90.6}	&\textbf{92.3} \\
 
\bottomrule
\end{NiceTabular}%
}
\vspace{-0.7cm}
\end{table}

\subsection{Evaluations}
To verify the effectiveness of the design choices of Gear-NeRF, we perform extensive ablation studies on the \emph{Truck} scene of the Google Immersive dataset~\cite{broxton2020immersive}.

\noindent\textbf{Motion-aware Temporal Sampling:} An intuitive strategy for adjusting the temporal sampling is to directly modify the temporal resolution of the 4D feature volume and make it uniform across all regions of the scene. In \Cref{tab:ablation}, we demonstrate the performance of this naive temporal sampling strategy using medium (25) or dense (100) temporal resolutions on 100-frame input videos. The results show that increasing the temporal resolution universally does not necessarily yield better performance. In contrast, our proposed method, which utilizes motion-aware temporal sampling, achieves the best results. We surmise that this may be attributed to the model distributing the volume's capacity sparsely across a large number of time steps, which is not the case for our method. 

\noindent\textbf{Motion-aware Spatial Sampling:} To validate the effectiveness of our motion-aware spatial sampling strategy, we compare it against other sampling strategies.
The first variant involves uniform sampling along the ray, and the second uses a sample prediction network (SPN) like Attal \etal \cite{attal2023hyperreel} for sampling. 
These variants sample 128 points on each ray. 
Results in \Cref{tab:ablation} show that our full model employing motion-aware spatial sampling outperforms these variants, perhaps by having a better sense of where to sample more from, derived from its semantic-aware embedding.

\noindent\textbf{SAM embedding:}
To verify that introducing SAM embedding can improve rendering quality (rather than just enabling segmentation or tracking), we tested a variant of Gear-NeRF without SAM embedding (it instead thresholds loss maps of rendered RGB frames to obtain upshift masks). 
As shown in \Cref{tab:ablation}, the absence of SAM embedding for guiding gear assignment reduces the model's rendering quality. 

\noindent\textbf{Number of Gears:} We ablate on the numbers of gears. Increasing the number of gears allows for more fine-grained motion-aware spatio-temporal sampling, while increasing the computational cost. As shown in \Cref{tab:ablation}, a choice of up to 4 gear levels seems optimal, further increasing the number of gears does not result in significant improvements.

\begin{table}[t]
\centering
\caption{\textbf{Ablation Study:} Ablations on our spatio-temporal sampling strategy and the number of gears (Truck/Google Immersive). \textbf{Best} and \underline{second best} results are highlighted. }
\vspace{-0.3cm}
\label{tab:ablation}
\resizebox{\linewidth}{!}{%
\setlength{\tabcolsep}{4pt}%
\begin{NiceTabular}{@{}lccc@{}}[code-before =%
\rectanglecolor{gray!20}{9-1}{9-5}%
]
\toprule
\thead[l]{Method} & \thead[c]{PSNR ($\uparrow$)} & \thead[c]{SSIM ($\uparrow$)} & \thead[c]{LPIPS ($\downarrow$)} \\ 
\midrule
Naive temporal sampling (medium) & 26.93 & 0.778 & 0.144 \\
Naive temporal sampling (dense) & 26.85 & 0.760 & 0.161 \\ \midrule
Naive spatial sampling & 24.80 & 0.734 & 0.222 \\
SPN spatial sampling & 26.54 & 0.787 & 0.162 \\ \midrule
Ours (w/o embedding) & 27.10 & 0.831 & 0.177 \\ \midrule
Ours ($N_{\text{gear}}=2$) & 26.93 & 0.785 & 0.166 \\
Ours ($N_{\text{gear}}=3$) & 27.14 & 0.875 & 0.145 \\
Ours ($N_{\text{gear}}=4$) & \textbf{27.49} & \ul{0.892} & \ul{0.136} \\
Ours ($N_{\text{gear}}=5$) & \ul{27.46} & \textbf{0.901} & \textbf{0.131} \\
\bottomrule
\end{NiceTabular}%
}
\vspace{-0.7cm}
\end{table}

\section{Conclusions}
In this work, we introduced \emph{Gear-NeRF}, an extension of dynamic NeRFs that leverages semantic information from powerful segmentation models for stratified modeling of dynamic scenes.
Our approach learns a 4D (spatio-temporal) semantic embedding and introduces the concept of ``gears'' for differentiated modeling of scene regions based on their motion intensity. 
With determined gear assignments, Gear-NeRF adaptively adjusts its spatio-temporal sampling resolution to improve the photo-realism of rendered views.
At the same time, Gear-NeRF provides the new functionality of free-viewpoint object tracking with user prompts as simple as a click.
Our empirical studies underscore the effectiveness of Gear-NeRF, showcasing state-of-the-art performance in both rendering quality and object tracking across multiple challenging datasets. 

\renewcommand\thefigure{\Alph{figure}} 
\renewcommand\thetable{\Alph{table}}   
\renewcommand\thesection{\Alph{section}}   
\setcounter{figure}{0} 
\setcounter{table}{0} 
\setcounter{section}{0} 

\newpage

\section{Appendix}
We begin this appendix by reporting \underline{per-scene rendering results} of Gear-NeRF compared to competing methods, both qualitatively and quantitatively. In \Cref{sec:add_track}, we present performance comparisons for the task of \underline{tracking in novel views}, a new contribution of this work, and compare against baselines adapted for this task. We then present  \underline{additional ablation studies}, discussing the sensitivity of our method to the choice of appropriate hyper-parameters in \Cref{sec:ablation_res}. 

\begin{figure*}[t]
\centering
\includegraphics[width=\linewidth]{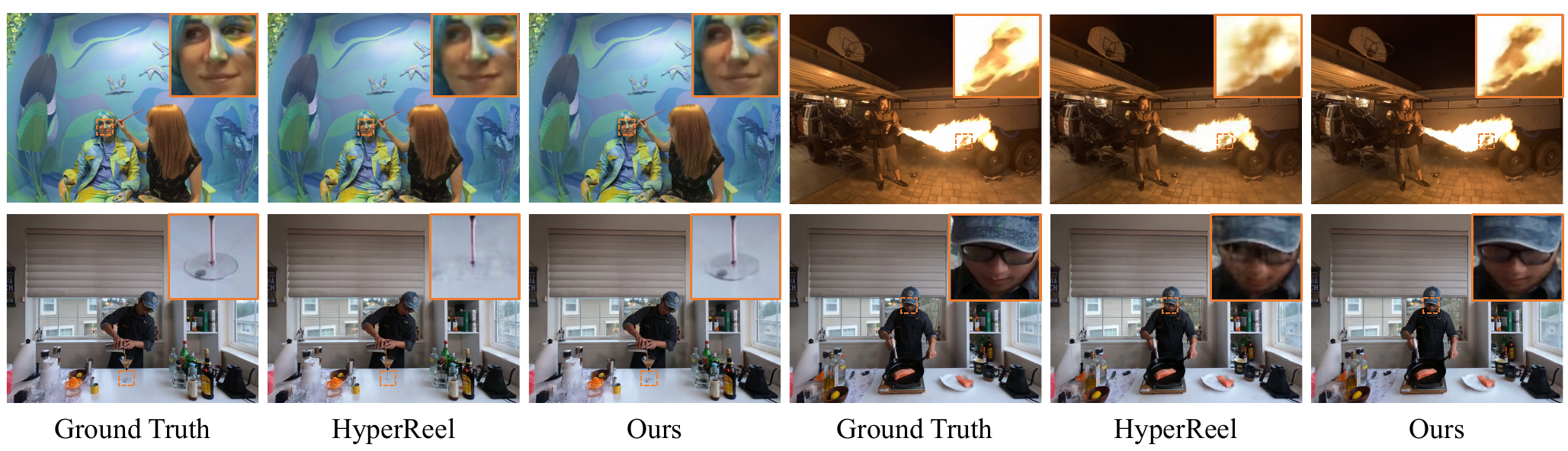}
\caption{\textbf{Qualitative comparisons of competing methods for the task of novel view synthesis of some additional dynamic scenes for the Google Immersive~\cite{broxton2020immersive} (top row) and the Neural 3D Video~\cite{li2022neural} (bottom row) datasets.} 
}
\label{fig:addn_render_result}
\end{figure*}

\begin{table}[b]
\centering
\caption{\textbf{Per-scene quantitative comparisons for the task of novel view synthesis for dynamic scenes on the Technicolor dataset~\cite{technicolor}.} \textbf{Best} and \underline{second best} results are highlighted.}
\label{tab:technicolor}
\resizebox{\linewidth}{!}{%
\setlength{\tabcolsep}{5pt}%
\begin{NiceTabular}{@{}llccc@{}}[code-before =%
\rectanglecolor{gray!20}{5-2}{5-5}%
\rectanglecolor{gray!20}{9-2}{9-5}%
\rectanglecolor{gray!20}{13-2}{13-5}%
\rectanglecolor{gray!20}{17-2}{17-5}%
]
\toprule
\thead[l]{Scene} & \thead[l]{Method} & \thead[c]{PSNR ($\uparrow$)} & \thead[c]{SSIM ($\uparrow$)} & \thead[c]{LPIPS ($\downarrow$)} \\ 
\midrule
\multirow{4}{*}{Train}
 & ST-NeRF~\cite{stnerf} & 29.16 & 0.877 & 0.070 \\
 & HyperReel~\cite{attal2023hyperreel} & \ul{29.18} & \ul{0.894} & \ul{0.054} \\
 & MixVoxels~\cite{wang2022mixed} & 27.34 & 0.830 & 0.058 \\
 & Ours & \textbf{30.55}& \textbf{0.957} & \textbf{0.049} \\
\midrule
\multirow{4}{*}{Theater}
 & ST-NeRF~\cite{stnerf} & 31.57 & 0.866 & 0.133 \\
 & HyperReel~\cite{attal2023hyperreel} & \ul{31.69} & 0.863 & \ul{0.131} \\
 & MixVoxels~\cite{wang2022mixed} & 27.34 & \textbf{0.888} & 0.134 \\
 & Ours & \textbf{32.56} & \ul{0.887} & \textbf{0.067} \\
\midrule
\multirow{4}{*}{Painter}
 & ST-NeRF~\cite{stnerf} & 35.14 & 0.911 & 0.102 \\
 & HyperReel~\cite{attal2023hyperreel} & \ul{35.38} &\ul{0.916} & 0.091 \\
 & MixVoxels~\cite{wang2022mixed} & 34.18 & 0.900 & \ul{0.076} \\
 & Ours & \textbf{36.35} & \textbf{0.928} & \textbf{0.073} \\
\midrule
\multirow{4}{*}{Birthday}
 & ST-NeRF~\cite{stnerf} & 27.55 & \ul{0.877} & 0.097 \\
 & HyperReel~\cite{attal2023hyperreel} & \ul{27.91} & 0.873 & \ul{0.090} \\
 & MixVoxels~\cite{wang2022mixed} & 27.11 & 0.749 & 0.142 \\
 & Ours & \textbf{29.38} & \textbf{0.904} & \textbf{0.041} \\
\bottomrule
\end{NiceTabular}%
}
\end{table}

\begin{table}[t]
\centering
\caption{\textbf{Per-scene quantitative comparisons for the task of novel view synthesis for dynamic scenes on the Google Immersive dataset~\cite{broxton2020immersive}.} \textbf{Best} and \underline{second best} results are highlighted.}
\label{tab:immersive}
\resizebox{0.95\linewidth}{!}{%
\setlength{\tabcolsep}{5pt}%
\begin{NiceTabular}{@{}llccc@{}}[code-before =%
\rectanglecolor{gray!20}{5-2}{5-5}%
\rectanglecolor{gray!20}{9-2}{9-5}%
\rectanglecolor{gray!20}{13-2}{13-5}%
\rectanglecolor{gray!20}{17-2}{17-5}%
\rectanglecolor{gray!20}{21-2}{21-5}%
\rectanglecolor{gray!20}{25-2}{25-5}%
\rectanglecolor{gray!20}{29-2}{29-5}%
\rectanglecolor{gray!20}{33-2}{33-5}%
\rectanglecolor{gray!20}{37-2}{37-5}%
]
\toprule
\thead[l]{Scene} & \thead[l]{Method} & \thead[c]{PSNR ($\uparrow$)} & \thead[c]{SSIM ($\uparrow$)} & \thead[c]{LPIPS ($\downarrow$)} \\ 
\midrule
\multirow{4}{*}{Flames}
 & HexPlane~\cite{cao2023hexplane} & 29.31 & 0.808 & 0.189 \\
 & HyperReel~\cite{attal2023hyperreel} & \ul{29.66} & \ul{0.895} & \ul{0.129} \\
 & MixVoxels~\cite{wang2022mixed} & 29.01 & 0.819 & 0.180 \\
 & Ours & \textbf{30.87} & \textbf{0.903} & \textbf{0.120} \\
\midrule
\multirow{4}{*}{Truck}
 & HexPlane~\cite{cao2023hexplane} & 26.89 & 0.819 & 0.161 \\
 & HyperReel~\cite{attal2023hyperreel} & \ul{27.20} & 0.850 & \ul{0.153} \\
 & MixVoxels~\cite{wang2022mixed} & 26.59 & \ul{0.877} & 0.194 \\
 & Ours & \textbf{27.46} & \textbf{0.892} & \textbf{0.136} \\
\midrule
\multirow{4}{*}{Horse}
 & HexPlane~\cite{cao2023hexplane} & 28.45 & 0.887 & 0.121 \\
 & HyperReel~\cite{attal2023hyperreel} & \ul{28.56} & \ul{0.892} & \ul{0.114} \\
 & MixVoxels~\cite{wang2022mixed} & 28.13 & 0.773 & 0.190 \\
 & Ours & \textbf{29.05} & \textbf{0.895} & \textbf{0.110} \\
\midrule
\multirow{4}{*}{Car}
 & HexPlane~\cite{cao2023hexplane} & 24.13 & 0.719 & 0.261 \\
 & HyperReel~\cite{attal2023hyperreel} & \ul{24.58} & \ul{0.740} & \ul{0.215} \\
 & MixVoxels~\cite{wang2022mixed} & 24.37 & 0.724 & 0.249 \\
 & Ours & \textbf{25.12} & \textbf{0.783} & \textbf{0.179} \\
\midrule
\multirow{4}{*}{Welder}
 & HexPlane~\cite{cao2023hexplane} & 25.89 & 0.778 & 0.250 \\
 & HyperReel~\cite{attal2023hyperreel} & \ul{26.07} & 0.793 & \ul{0.220} \\
 & MixVoxels~\cite{wang2022mixed} & 24.59 & \textbf{0.818} & 0.277 \\
 & Ours & \textbf{26.36} & \ul{0.810} & \textbf{0.187} \\
\midrule
\multirow{4}{*}{Exhibit}
 & HexPlane~\cite{cao2023hexplane} & 29.93 & 0.874 & 0.159 \\
 & HyperReel~\cite{attal2023hyperreel} & \ul{31.53} & 0.907 & \ul{0.090} \\
 & MixVoxels~\cite{wang2022mixed} & 28.35 & \ul{0.915} & 0.148 \\
 & Ours & \textbf{31.73} & \textbf{0.920} & \textbf{0.064} \\
\midrule
\multirow{4}{*}{Face Paint 1}
 & HexPlane~\cite{cao2023hexplane} & 28.48 & 0.841 & 0.169 \\
 & HyperReel~\cite{attal2023hyperreel} & \textbf{29.83} & \textbf{0.922} & \ul{0.093} \\
 & MixVoxels~\cite{wang2022mixed} & 27.84 & 0.847 & 0.185 \\
 & Ours & \ul{29.15} &     \ul{0.901} & \textbf{0.082} \\
\midrule
\multirow{4}{*}{Face Paint 2}
 & HexPlane~\cite{cao2023hexplane} & 28.58 & 0.833 & 0.148 \\
 & HyperReel~\cite{attal2023hyperreel} & \ul{28.94} & \ul{0.893} & \ul{0.106} \\
 & MixVoxels~\cite{wang2022mixed} & 27.50 & 0.849 & 0.231 \\
 & Ours & \textbf{29.24} & \textbf{0.903} & \textbf{0.076} \\
\midrule
\multirow{4}{*}{Cave}
 & HexPlane~\cite{cao2023hexplane} & 27.35 & 0.715 & 0.231 \\
 & HyperReel~\cite{attal2023hyperreel} & \ul{28.48} & \ul{0.867} & \ul{0.184} \\
 & MixVoxels~\cite{wang2022mixed} & 27.93 & 0.894 & 0.224 \\
 & Ours & \textbf{29.68} & \textbf{0.880} & \textbf{0.144} \\
\bottomrule
\end{NiceTabular}%
}
\end{table}

\subsection{Per-Scene Rendering Results}
\label{sec:add_ren}

In this section, we present a quantitative evaluation of Gear-NeRF and competing techniques for the task of rendering dynamic scenes from novel views, on a per-scene basis for each of the three datasets we conduct experiments on: (i) The Technicolor Lightfield Dataset~\cite{technicolor} (ii) The Neural 3D Video Dataset~\cite{li2022neural}, and the (iii) The Google Immersive Dataset~\cite{broxton2020immersive}. Moreover, to further demonstrate the generalizability of our method vis-\'a-vis our closest competing baseline, HyperReel~\cite{attal2023hyperreel}, we report its performance versus that of our method on some additional sequences for each of these three datasets. 

\Cref{tab:technicolor}, \Cref{tab:immersive}, and \Cref{tab:n3d} show per-scene quantitative comparison results of our approach against competing methods on the Technicolor dataset \cite{technicolor}, the Neural 3D Video dataset \cite{li2022neural}, and the Google Immersive dataset \cite{broxton2020immersive}, respectively. 
The averaged results are presented in Table 1 of the paper and are derived from these per-scene results. We see that in all but a couple of sequences (``Cut Roasted Beef'' from the Neural 3D video dataset oe ``Theater'' from the Technicolor dataset) our proposed approach outperforms all other competing methods, across all the metrics, attesting to the effectiveness of our method. 
Even under occasional circumstances when that is not the case, our method still reports performance comparable to HyperReel. \Cref{fig:addn_render_result} presents qualitative comparisons of rendering results, by our method and HyperReel for some sequences from the Google Immersive~\cite{broxton2020immersive} and the Neural 3D Video~\cite{li2022neural} datasets. As is evident from the figure, the frames synthesized by our method look less blurry and better preserves the details (for instance the eye of the lady, the flame, the stem of the glass, or the glasses of the man with the hat) which underscores the effectiveness of our method. More qualitative results can be seen in the \ul{attached video}.

\begin{table}[t]
\centering
\caption{\textbf{Per-scene quantitative comparisons for the task of novel view synthesis for dynamic scenes on the Neural 3D Video~\cite{li2022neural}.} \textbf{Best} and \underline{second best} results are highlighted.}
\label{tab:n3d}
\resizebox{\linewidth}{!}{%
\setlength{\tabcolsep}{5pt}%
\begin{NiceTabular}{@{}llccc@{}}[code-before =%
\rectanglecolor{gray!20}{5-2}{5-5}%
\rectanglecolor{gray!20}{9-2}{9-5}%
\rectanglecolor{gray!20}{13-2}{13-5}%
\rectanglecolor{gray!20}{17-2}{17-5}%
\rectanglecolor{gray!20}{21-2}{21-5}%
\rectanglecolor{gray!20}{25-2}{25-5}%
]
\toprule
\thead[l]{Scene} & \thead[l]{Method} & \thead[c]{PSNR ($\uparrow$)} & \thead[c]{SSIM ($\uparrow$)} & \thead[c]{LPIPS ($\downarrow$)} \\
\midrule
\multirow{4}{*}{Cut Roasted Beef}
& ST-NeRF~\cite{stnerf} & \textbf{32.97} & 0.950 & 0.047 \\
& HyperReel~\cite{attal2023hyperreel} & 32.63 & 0.942 & \textbf{0.049} \\
& MixVoxels~\cite{wang2022mixed} & 32.34 & \textbf{0.962} & 0.138 \\
& Ours & \ul{32.74} & \ul{0.944} & \ul{0.057} \\
\midrule
\multirow{4}{*}{Coffee Martini}
& ST-NeRF~\cite{stnerf} & \ul{29.18} & \ul{0.904} & 0.102 \\
& HyperReel~\cite{attal2023hyperreel} & 28.43 & 0.896 & 0.090 \\
& MixVoxels~\cite{wang2022mixed} & 28.08 & 0.901 & \ul{0.079} \\
& Ours & \textbf{29.71} & \textbf{0.918} & \textbf{0.070} \\
\midrule
\multirow{4}{*}{Flame Steak}
& ST-NeRF~\cite{stnerf} & 31.75 & 0.903 & 0.061 \\
& HyperReel~\cite{attal2023hyperreel} & \ul{32.49} & \ul{0.946} & \ul{0.051} \\
& MixVoxels~\cite{wang2022mixed} & 31.54 & \ul{0.946} & 0.133 \\
& Ours & \textbf{33.20} & \textbf{0.952} & \textbf{0.045} \\
\midrule
\multirow{4}{*}{Cook Spinach}
& ST-NeRF~\cite{stnerf} & \ul{32.84} & 0.942 & \ul{0.049} \\
& HyperReel~\cite{attal2023hyperreel} & 32.56 & 0.940 & 0.056 \\
& MixVoxels~\cite{wang2022mixed} & 31.71 & \textbf{0.960} & 0.144 \\
& Ours & \textbf{33.18} & \ul{0.946} & \textbf{0.046} \\
\midrule
\multirow{4}{*}{Flame Salmon}
& ST-NeRF~\cite{stnerf} & 27.74 & 0.781 & 0.132 \\
& HyperReel~\cite{attal2023hyperreel} & 28.03 & 0.891 & \ul{0.100} \\
& MixVoxels~\cite{wang2022mixed} & \ul{28.88} & \textbf{0.930} & 0.212 \\
& Ours & \textbf{29.66} & \ul{0.912} & \textbf{0.073} \\
\midrule
\multirow{4}{*}{Sear Steak}
& ST-NeRF~\cite{stnerf} & 31.72 & 0.862 & 0.094 \\
& HyperReel~\cite{attal2023hyperreel} & \textbf{32.58} & \ul{0.951} & \textbf{0.046} \\
& MixVoxels~\cite{wang2022mixed} & 31.60 & \textbf{0.967} & 0.128 \\
& Ours & \ul{32.31} & 0.942 & \ul{0.054} \\
\bottomrule
\end{NiceTabular}%
}
\end{table}

\begin{figure}[t]
\centering
\includegraphics[width=\linewidth]{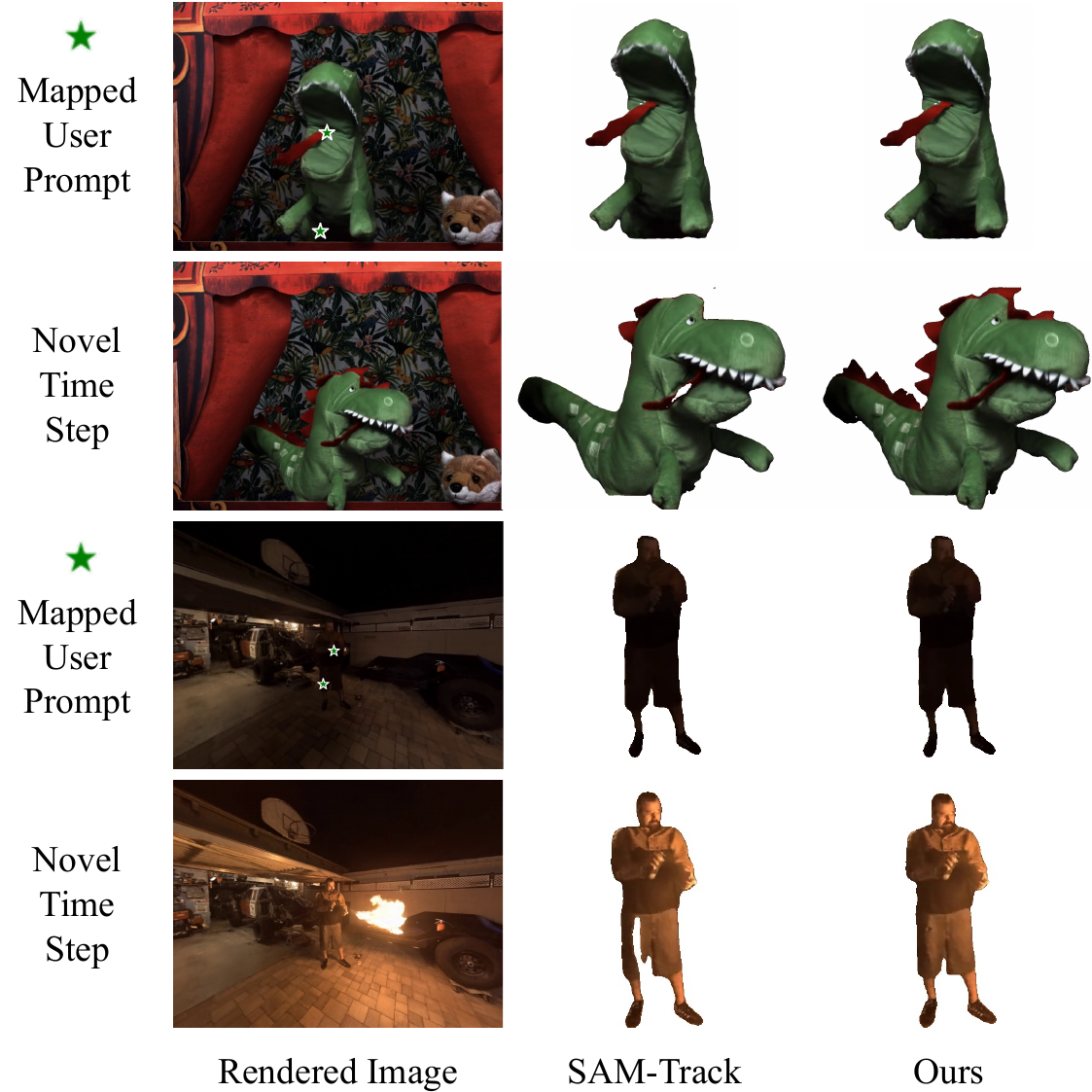}
\caption{\textbf{Qualitative comparisons of click-based novel-view tracking of our method versus SAM-Track~\cite{cheng2023segment}.} }
\label{fig:add_track}
\end{figure}

\begin{figure}[t]
\centering
\includegraphics[width=\linewidth]{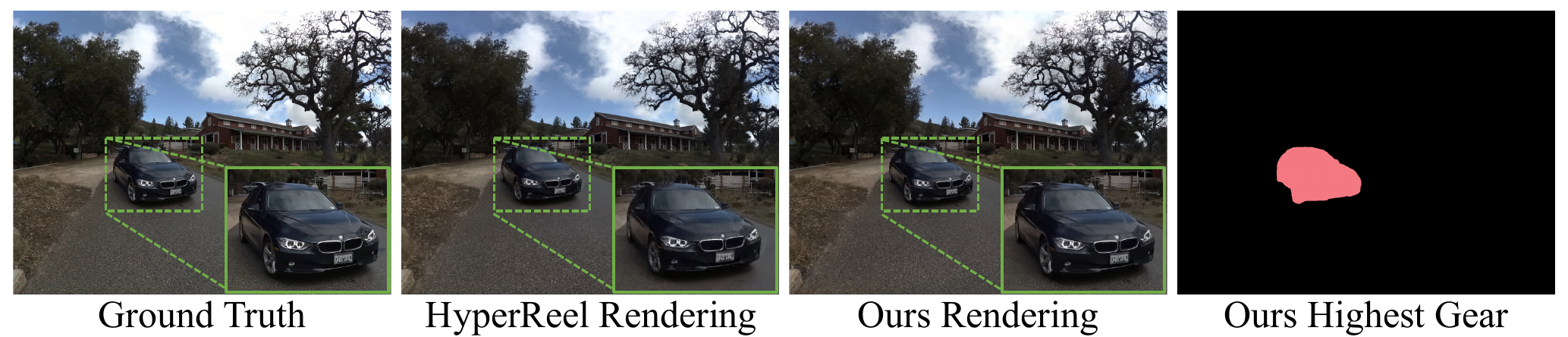}
\caption{\textbf{Gear selection and rendering of non-Lambertian objects.}
}
\label{fig:car}
\vspace{-0.4cm}
\end{figure}

\noindent\textbf{Non-Lambertian Surfaces:}
%
While non-Lambertian surfaces are known to pose challenges for rendering, we observe that they don't undermine the gear selection, perhaps due to object priors from SAM. E.g. the \textit{car} in the scene in \Cref{fig:car} includes reflective surfaces, like windshield yet it is assigned the right gear with its details better reconstructed than competing methods. 

\subsection{Novel-View Tracking Results}
\label{sec:add_track}

Being the first method to achieve free-viewpoint tracking of target objects in the NeRF setting, our approach does not have direct baselines, to the best of our knowledge. Hence, we use the following as baselines for benchmarking: (i) The static scene segmentation approach, SA3D \cite{sa3d} mentioned in Section 5 of the main paper.  (ii) We also compare against a monocular video tracking baseline called SAM-Track \cite{cheng2023segment} -- a method based on SAM \cite{sam} for object tracking in monocular videos.
Since SAM-Track only takes a monocular video as input and does not consider the 3D information, we adopted the following procedure to use it as a baseline: Given user-provided click(s) in an input view, we utilize our radiance field representation to map these clicks to a desired target/novel view. SAM-Track can then be used to perform object tracking in the target view using the mapped click(s) as prompts.
As the quantitative results in \Cref{tab:sam_track} indicate, our method outperforms SAM-Track across all metrics on all datasets for the task of desired novel/target view object tracking. This may be attributed to our method's capability of learning the semantics of the scene by leveraging the 4D SAM embedding field. A rendered SAM feature map is fed into the SAM decoder to obtain the mask of the target object at every time step. In contrast, SAM-Track uses SAM to acquire the object mask only for the first frame and employs a mask tracker~\cite{yang2022decoupling} to obtain masks for subsequent time steps. This is also demonstrated in \Cref{fig:add_track} where our approach better renders the scene without introducing artifacts as opposed to SAM-Track. More qualitative results can be seen in the \ul{attached video}.

\Cref{tab:tracking_supp} reveals that our approach better segments the target object, given a rendered frame, as compared to SA3D~\cite{sa3d}. We attribute this gain to the fact that our method unlike SA3D reasons about the temporal dynamics of the scene and can thus better assess/predict the location of the target object.

\begin{table}[t]
\centering
\caption{\textbf{Quantitative comparisons for fixed novel view tracking versus SAM-Track~\cite{cheng2023segment}.}}
\label{tab:sam_track}
\resizebox{0.9\linewidth}{!}{%
\setlength{\tabcolsep}{5pt}%
\begin{NiceTabular}{@{}llcc@{}}[code-before =%
\rectanglecolor{gray!20}{3-2}{3-5}%
\rectanglecolor{gray!20}{5-2}{5-5}%
\rectanglecolor{gray!20}{7-2}{7-5}%
]
\toprule
\thead[l]{Dataset} & \thead[l]{Method} & \thead[c]{mIoU} & \thead[c]{Accuracy} \\ 
\midrule
\multirow{2}{*}{Technicolor~\cite{technicolor}}
& SAM-Track~\cite{cheng2023segment} & 95.6 & 96.1 \\
 & Ours & \textbf{96.0} & \textbf{96.9} \\
\midrule
\multirow{2}{*}{Neural 3D Video~\cite{li2022neural}}
 & SAM-Track~\cite{cheng2023segment} & 94.1 & 94.5 \\
 & Ours & \textbf{95.1} & \textbf{95.5} \\
\midrule
\multirow{2}{*}{Google Immersive~\cite{broxton2020immersive}}
 & SAM-Track~\cite{cheng2023segment} & 93.4 & 94.0 \\
 & Ours & \textbf{95.7} & \textbf{96.3} \\
\bottomrule
\end{NiceTabular}%
}
\end{table}

\begin{table}[t]
\centering
\caption{\textbf{Quantitative comparisons for free-viewpoint tracking:} t-mIoU and t-Acc are metrics used for evaluating novel view masks at novel time steps, not applicable to SA3D. Reported metrics are averages over all scenes for each dataset.}
\label{tab:tracking_supp}
\resizebox{1\linewidth}{!}{%
\setlength{\tabcolsep}{5pt}%
\begin{NiceTabular}{@{}llcccc@{}}[code-before =%
\rectanglecolor{gray!20}{3-2}{3-6}%
\rectanglecolor{gray!20}{5-2}{5-6}%
\rectanglecolor{gray!20}{7-2}{7-6}%
]
\toprule
\thead[l]{Dataset} & \thead[l]{Method} & \thead[c]{mIoU ($\uparrow$)} & \thead[c]{Acc. ($\uparrow$)} & \thead[c]{t-mIoU ($\uparrow$)} & \thead[c]{t-Acc. ($\uparrow$)} \\ 
\midrule
\multirow{2}{*}{Technicolor~\cite{technicolor}}
 & SA3D~\cite{sa3d} & 96.4	& 97.1& N/A &N/A \\
 & Ours &\textbf{ 97.4}	 &\textbf{97.6} &	\textbf{92.1} &	\textbf{93.3 }\\
\midrule
\multirow{2}{*}{Google Immersive~\cite{broxton2020immersive}}
 & SA3D~\cite{sa3d} &94.1	&94.8	&N/A	&N/A\\
 & Ours & \textbf{94.3}&	\textbf{95.0}&	\textbf{91.5}&	\textbf{92.8} \\
 \midrule
\multirow{2}{*}{Neural 3D Video~\cite{li2022neural}}
 & SA3D~\cite{sa3d} & 93.1 &	94.0 &	N/A &	N/A\\
 & Ours &\textbf{ 93.4}	&\textbf{94.3}	&\textbf{90.6}	&\textbf{92.3} \\
 
\bottomrule
\end{NiceTabular}%
}
\end{table}

\subsection{Additional Ablation Studies}
\label{sec:ablation_res}

In this section, we present some additional ablation results on the hyper-parameters of our model.

\noindent\textbf{Top-$k$ in Gear Assignment Updates:} For gear assignment updates, we employ a patch-based approach to identify regions with the top-$k$ highest or lowest average rendering loss to obtain positive or negative prompts for subsequent steps. We perform an ablation study on the \emph{Truck} scene of the Google Immersive dataset~\cite{broxton2020immersive}.
\Cref{tab:ablation_topk} reveals that both excessively high or low values of $k$ do not yield optimal performance. We note that a selection of $k=4$ or $5$ leads to gear upshifts for inappropriate regions, weakening the efficacy of our motion-aware spatio-temporal sampling strategy. 
In our experiments, we uniformly applied $k=3$ across all scenes, which yielded satisfactory results.

\begin{table}[t]
\centering
\caption{\textbf{Ablation study on the top-$k$ selection in gear assignment.} \textbf{Best} and \underline{second best} results are highlighted.}
\label{tab:ablation_topk}
\resizebox{0.7\linewidth}{!}{%
\setlength{\tabcolsep}{4pt}%
\begin{NiceTabular}{@{}lccc@{}}
\toprule
\thead[l]{Method} & \thead[c]{PSNR ($\uparrow$)} & \thead[c]{SSIM ($\uparrow$)} & \thead[c]{LPIPS ($\downarrow$)} \\ 
\midrule
Ours ($k=1$) & 27.10 & 0.879 & \ul{0.139} \\
Ours ($k=2$) & \ul{27.43} & \ul{0.890} & 0.145 \\
Ours ($k=3$) & \textbf{27.49} & \textbf{0.892} & \textbf{0.136} \\
Ours ($k=4$) & 26.14 & 0.777 & 0.158 \\
Ours ($k=5$) & 26.39 & 0.790 & 0.161 \\
\bottomrule
\end{NiceTabular}%
}
\end{table}

\begin{table}[t]
\centering
\caption{\textbf{Ablation Study on the point splitting strategy in motion-aware spatial sampling.} \textbf{Best} and \underline{second best} results are highlighted.}
\label{tab:ablation_point_splitting}
\resizebox{0.9\linewidth}{!}{%
\setlength{\tabcolsep}{4pt}%
\begin{NiceTabular}{@{}lccc@{}}
\toprule
\thead[l]{Method} & \thead[c]{PSNR ($\uparrow$)} & \thead[c]{SSIM ($\uparrow$)} & \thead[c]{LPIPS ($\downarrow$)} \\ 
\midrule
Ours ($2^{p(\mathbf{x}, t)-1}$) & \ul{27.49} & \ul{0.892} & \ul{0.136} \\
Ours ($3^{p(\mathbf{x}, t)-1}$) & \textbf{27.98} & \textbf{0.914} & \textbf{0.125} \\
Ours ($2p(\mathbf{x}, t) - 1$) & 26.46 & 0.815 & 0.140 \\
\bottomrule
\end{NiceTabular}%
}
\end{table}

\noindent\textbf{Sampling Point Splitting:} In our motion-aware spatial sampling, we adopt a 3D sampling point-splitting strategy. Specifically, we split each sampled 3D point into $2^{p(\mathbf{x}, t)-1}$ points.
We conduct an ablation study on the number of points a sampling point is split into. To elaborate, in addition to splitting one point into $2^{p(\mathbf{x}, t)-1}$ points, we explore variants, including splitting into $3^{p(\mathbf{x}, t)-1}$ points and $2p(\mathbf{x}, t) - 1$ points, on the \emph{Truck} scene of the Google Immersive dataset~\cite{broxton2020immersive}.
As shown in \Cref{tab:ablation_point_splitting}, the additional sampling points generated by the $2p(\mathbf{x}, t) - 1$ strategy are insufficient, resulting in a decrease in rendering quality. In contrast, $3^{p(\mathbf{x}, t)-1}$ achieves better quality than $2^{p(\mathbf{x}, t)-1}$. However, an excessive number of sampling points leads to a reduction in training speed, while providing a marginal performance boost, which is why we stick with the strategy of splitting into $2^{p(\mathbf{x}, t)-1}$ points.

{
    \small
    \bibliographystyle{ieeenat_fullname}
    \bibliography{main}
}

\end{document}

%% file: preamble.tex
\usepackage[dvipsnames]{xcolor}

\usepackage{breqn}
\usepackage{booktabs}
\usepackage{makecell}
\usepackage{multirow}
\usepackage{nicematrix}
\usepackage{multirow}
\usepackage{makecell}
\usepackage{array}

\newcolumntype{L}[1]{>{\raggedright\let\newline\\\arraybackslash\hspace{0pt}}m{#1}}
\newcolumntype{C}[1]{>{\centering\let\newline\\\arraybackslash\hspace{0pt}}m{#1}}
\newcolumntype{R}[1]{>{\raggedleft\let\newline\\\arraybackslash\hspace{0pt}}m{#1}}

\usepackage{soul}

%% file: definitions.tex
\newcommand{\set}[1]{\left\{#1\right\}}

\newcommand{\vid}{V}

\renewcommand{\set}[1]{\left\{#1\right\}}